\documentclass{llncs}
\usepackage{makeidx}  % allows for indexgeneration
\usepackage{amsmath}
\usepackage{amssymb}
\usepackage[nodisplayskipstretch]{setspace}
\usepackage{subcaption}
\usepackage{multirow}
\usepackage{graphicx}
\usepackage{tabularx}
\graphicspath{{figs/}}

\usepackage{booktabs}
\usepackage[usenames, dvipsnames]{color}

\newcommand{\matr}[1]{\mathbf{#1}} % undergraduate algebra version

\let\subparagraph\paragraph

\usepackage{titlesec}
\titlespacing\section{0pt}{12pt plus 4pt minus 2pt}{6pt plus 2pt minus 2pt}
\titlespacing\subsection{0pt}{12pt plus 2pt minus 2pt}{0pt plus 2pt minus 2pt}
\titlespacing\subsubsection{0pt}{6pt plus 4pt minus 2pt}{0pt plus 2pt minus 2pt}
\usepackage[hidelinks]{hyperref}

\newcolumntype{C}{>{\centering\arraybackslash}X}

\begin{document}
\frontmatter          % for the preliminaries
\pagestyle{headings}  % switches on printing of running heads
\addtocmark{Hamiltonian Mechanics} % additional mark in the TOC
\title{A Neural Attention Model for Categorizing Patient Safety Events}
\author{Arman Cohan\inst{1} \and Allan Fong\inst{2} \and Nazli Goharian\inst{1} \and Raj Ratwani\inst{2}}
\authorrunning{Arman Cohan et al.} % abbreviated author list (for running head)
\tocauthor{Arman Cohan, Allan Fong, Nazli Goharian, and Raj Ratwani}
\institute{Department of Computer Science, Georgetown University, Washington DC, USA\\
\email{\{arman,nazli\}@ir.cs.georgetown.edu}
\and
National Center for Human Factors in Healthcare, MedStar Health\\
Washington DC, USA\\
\email{\{allan.fong,raj.ratwani\}@medicalhfe.org}}
\titlerunning{Hamiltonian Mechanics}  % abbreviated title (for running head)
% \author{No Author Given\inst{1} \and Roger Temam\inst{2}
% Jeffrey Dean \and David Grove \and Craig Chambers \and Kim~B.~Bruce \and
% Elsa Bertino}
% %
% \authorrunning{Ivar Ekeland et al.} % abbreviated author list (for running head)
% %
% %%%% list of authors for the TOC (use if author list has to be modified)
% \tocauthor{Ivar Ekeland, Roger Temam, Jeffrey Dean, David Grove,
% Craig Chambers, Kim B. Bruce, and Elisa Bertino}
% %
% \institute{Princeton University, Princeton NJ 08544, USA,\\
% \email{I.Ekeland@princeton.edu},\\ WWW home page:
% \texttt{http://users/\homedir iekeland/web/welcome.html}
% \and
% Universit\'{e} de Paris-Sud,
% Laboratoire d'Analyse Num\'{e}rique, B\^{a}timent 425,\\
% F-91405 Orsay Cedex, France}

\maketitle              % typeset the title of the contribution

\begin{abstract}
Medical errors are leading causes of death in the US and as such, prevention of these errors is paramount to promoting healthcare.
Patient Safety Event reports are narratives describing potential adverse events to the patients and are important in identifying, and preventing medical errors. We present a neural network architecture for identifying the type of safety events which is the first step in understanding these narratives. Our proposed model is based on a soft neural attention model to improve the effectiveness of encoding long sequences. Empirical results on two large-scale real-world datasets of patient safety reports demonstrate the effectiveness of our method with significant improvements over existing methods.
\keywords{Natural Language Processing, Text Categorization, Medical Text Processing, Deep Learning}
\end{abstract}
\section{Introduction}
There is an increasing demand for use of textual electronic health records and clinical notes to promote healthcare, and as such,
In recent years NLP/IR have become increasingly important in understanding, searching, and analyzing medical information \cite{yates2015extracting}.
%While healthcare providers are constantly improving their health services,
Human or system errors do occur frequently in the health centers, many of which can lead to serious harm to individuals. There are in fact an alarming number of annual death incidents (up to 200K) being reported due to medical errors \cite{american2013fast}; medical errors are shown to be the third leading cause of death in the US \cite{makary2016medical}. Many healthcare centers have deployed patient safety event reporting systems to better identify, mitigate, and prevent errors \cite{clarke2006system}. Patient safety event reports are narratives describing a safety event and they belong to different safety categories such as ``medication'', ``diagnosis'', ``treatment'', ``lab'', etc. Recently, due to the importance of patient safety reports, more healthcare centers are enforcing patient safety reporting, resulting in an overwhelming number of daily produced reports. Manual processing of all these reports to identify important cases, trends, or system issues is extremely difficult, inefficient, and expensive. The first step in understanding and analyzing these events is to identify their general categories. This task is challenging because the event descriptions can be very complex; the frontline staff usually focus more on taking care of the patient at the moment than to think through the classification schema when they later write a safety report. For example, an event where a patient fell after being given an incorrect medication might have been classified as ``\textsc{fall}'' however, the fall could be due to a mis-medication and therefore belong to the ``\textsc{medication}'' safety event. Without the ability to correctly identify the medication category, such problems will not be addressed. Therefore, classifying the patient safety reports not only helps in further search and analytic tasks, but also it contributes to reducing the human reporting errors.
% Given a safety report narrative, if we prompt the front line nurse or clinician with an appropriate event type, we can prevent or reduce the number of wrong event type indications.

In this paper, we present a method for categorizing the Patient Safety Reports as the first step towards understanding adverse events and the way to prevent them. Traditional approaches of text categorization rely on sparse feature extraction from clinical narratives and then classifying the types of events based on these feature representations. In these conventional methods, complex lexical relations and long-term dependencies of the narratives are not captured. We propose a neural attention architecture for classifying safety events, which performs the feature extraction, and type classification jointly; our proposed architecture is based on a combination of Convolutional Neural Networks (CNNs) and Recurrent Neural Networks (RNNs) with soft attention mechanism. We evaluate our method on two large scale datasets obtained from two large healthcare providers. We demonstrate that our proposed method significantly improves over several traditional baselines, as well as more recent neural network based methods.
% Specifically, our contributions are the following: (\textit{i}) A neural network based

\section{The proposed Neural Attention Architecture}

Our proposed model for classifying patient safety reports is a neural architecture based on Convolutional Neural Networks (CNN) and Recurrent Neural Networks (RNN) utilizing a soft attention mechanism. Our architecture is partially similar to models by \cite{kim:2014,kalchbrenner2014convolutional} in convolutional layers, to \cite{tang2015document} in recurrent layer, and to \cite{yang2016hierarchical} in the document modeling. Our point of departure is that unlike these works which are mainly targeted for sentence and short documents, we utilize a soft neural attention mechanism coupled with CNN and RNN to capture the more salient local features in longer sequences. Below we present the building blocks of our proposed architecture from bottom to the top.

\textbf{Embedding layer. }
Given a sequence of words $S=\langle w_1;w_2; ...;w_n\rangle$ where $w_i$ are words in the sequence and ``$:$'' is the concatenation operation, the embedding layer represents $S$ as an input vector $\matr{x}\in\mathbb{R}^{(m,d)}$ where $d$ is the embedding dimension size and $m$ is the maximum sequence length. $x_i$'s can be either initialized randomly or by pre-trained word embeddings, and then they can be jointly trained with the model.
% Represents a sequence of words $S=\langle w_1;w_2; ...;w_n\rangle$ with an input matrix $\matr{x}\in\mathbb{R}^{(n,d)}$ that can be either initialized randomly or by pre-trained word embeddings, and then can be jointly trained with the model.

\textbf{CNN. }
CNNs are feed-forward networks which include two main operations: \textit{convolution} and \textit{pooling}. Convolution is an operation on two functions (input and kernel) of real valued arguments \cite{lecun1998}.
%This operation expresses the amount of overlap of the kernel (filter) function $w$ as it is shifted over the input function $x$, and is defined as:
% \begin{equation}
% (x*w)(t)= \sum\limits_{a=-\infty}^{\infty} x(a)w(t-a)
% \end{equation}
In our context, in layer $\ell$ in the network, convolution operates on sliding windows of width $k_\ell$ on the input $\matr{x}_{\ell-1}$ and yields a feature map $F_\ell$:
\begin{equation}
  F_\ell^{(i)} = g(\matr{W}_\ell \;.\; \matr{x}^{(i,k_\ell)}_{\ell-1} + \matr{b}_\ell)
\end{equation}
\noindent where $\matr{W}_\ell$ and $\matr{b}_\ell$ are the shared wights and biases in layer $\ell$, $g$ is an activation function, and $\matr{x}^{(i,k_\ell)}=\langle x^{i-\frac{(k_\ell - 1)}{2}};...;x^{i+\frac{(k_\ell-1)}{2}}\rangle$ shows the sliding window of size $k_\ell$ centered at position $i$ on the input.
%The role of activation function is to transform or squash the output.
We use ReLU \cite{dahl2013improving} for the activation function (In our experiments ReLU showed the best results among other activation functions). For pooling, we use ``max-pooling'' operation whose role is to down-sample the feature map and capture significant local features. Similar to \cite{kim:2014}, we use filters of sizes from 2 to 6 to capture local features of different granularities. The convolution layer allows the model to learn the salient features that are needed for identifying the type of the safety events.

% In addition to convolutional layers just described, CNNs also contain pooling layers. Pooling is an operation performed on the resulting feature map from the convolution layer to obtain a condensed representation. We use the  CNNs are effective in capturing the regional important features using convolution and pooling operations \cite{kim:2014}.

\textbf{RNN. }
Unlike CNNs which are local feature encoders, RNNs can encode large windows of local features and capture long temporal dependencies. Given an input sequence $\matr{h}=(x_1, ..., x_T)$ where each $x_t\in\mathbb{R}^d$ is an input word vector of dimension $d$ at time step $t$, an RNN computes the hidden states $\matr{h}=(h_1, ..., h_T)$ and outputs $\matr{y}=(y_1, ..., y_T)$ according to the following equations \cite{elman1990finding}:
\small
\begin{align}
  h_t= & g(W^{(hh)}h_{t-1} + W^{(xh)}x_t + b_h)  &
  y_t= & W^{(hy)}h_{t} + b_y \label{eq-rnn}
\end{align}

\normalsize
\noindent where $W$ shows the weight matrices for the corresponding input,
%(e.g. $W^{(xh)}$ shows input-hidden weight matrix),
$b$ denotes the biases, and $g$ is the activation function. RNNs in theory, can capture temporal dependencies of any length. However, training RNNs in their basic form is problematic due to the \textit{vanishing gradient} problem \cite{pascanu2013difficulty}. Long Short-Term Memory (LSTM) \cite{hochreiter1997long} is a type of RNN that has several gates controlling the flow of information to be preserved or forgotten, and mitigates the vanishing gradient problem. We use the LSTM formulation as in \cite{graves2014towards}. We aslo employ bidirectional LSTM to capture both forward and backward temporal dependencies. Using this layer, we capture the dependencies between local features along long sequences.

% \begin{align*}

% \small
% \begin{subequations}
% \setstretch{0.7}
% \begin{align}
%   i_t &= \sigma(W^{(xi)}x_t+W^{(hi)}h_{t-1}+W^{(ci)}c_{t-1}+b_i)  \\
%   f_t &= \sigma(W^{(xf)}x_t+W^{(hf)}h_{t-1}+W^{(cf)}c_{t-1}+b_f) \\
%   c_t &= f_tc_{t-1}+i_t\; \tanh(W^{(xc)}x_t+W^{(hc)}h_{t-1}+b_c) \\
%   o_t &= \sigma(W^{(xo)}x_t + W^{(ho)}h_{t-1}+W^{(co)}c_t+b_o) \\
%   h_t &=o_t\; \tanh(c_t) \label{eq-lstm-out}
% \end{align}
% \end{subequations}
% \normalsize
% % \end{align}
% \noindent where $\sigma$ is the sigmoid function, $i$, $f$, $o$, and $c$ are the \textit{input gate}, \textit{forget gate}, \textit{ouput gate}, and \textit{cell state} vectors, respectively. All these vectors are at the same size as the hidden vector $h$.

% \begin{figure}
%   \includegraphics{neural-architecture}
%   \caption{The overall architecture of the model which consists of an embedding layer representing the input, convolution and pooling layers, recurrent layer and a softmax classifier.}
%   \label{fig:arch}
% \end{figure}

 % Figure \ref{fig:arch} shows the overall architecture of the model for classifying the patient safety reports.

\textbf{Neural attention.}
%The CNN layer captures the local features and RNN can capture long dependencies.
The trouble with RNNs for classification is that they encode the entire sequence into the vector at the last temporal step. While the application of RNNs have been successful in encoding sentences or short documents, in longer documents this can result in loss of information \cite{cho2014properties}, and putting more focus on the recent temporal entries \cite{sutskever2014sequence}. Bidirectional RNNs try to alleviate this problem by considering both the forward and backward context vectors. However, they suffer from the same problem in long sequences.

Inspired by work in machine-translation, to address this problem, we utilize the soft attention mechanism \cite{bahdanau2014neural}. Neural attention allows the model to decide which parts of the sequence are more important instead of directly considering the context vector output by the RNN. Specifically, instead of considering the final cell state of LSTM for the classification, we allow the model to attend to the important timesteps and build a context vector $c$ as follows:
\begin{equation}
  c = \sum\nolimits_{t=1}^T \alpha_{t}h_{t}
\end{equation}
where $\alpha_{t}$ are weights computed at each timestep $t$ for the state $h_{t}$ and are computed as follows:

\noindent\begin{minipage}{.5\linewidth}
\begin{equation}
  \alpha_{t} = \frac{\exp(e_{t}^\top z)}{\sum_{k=1}^T \exp(e_{k}^\top z)}
\end{equation}
\end{minipage}%
\begin{minipage}{.5\linewidth}
\begin{equation}
  e_{t} = f_{\mathrm{\textsc{att}}}( h_{t})
\end{equation}
\end{minipage}

\noindent where $f_{\mathrm{\textsc{att}}}$ is a function whose role is to capture the importance of $h_{t_i}$ and $z$ is a context vector that is learned jointly during training. We use a feed-forward network with ``$\tanh$'' activation function for $f_{\mathrm{\textsc{att}}}$. The context vector $c$ is then fed to a fully-connected and then a softmax layer to perform final classification.

\begin{table}[t]
  \renewcommand{\arraystretch}{0.7}
  \scriptsize
  \setlength{\tabcolsep}{1.2pt}
  \caption{Dataset characteristics}
  \label{tab:data1}
\begin{subtable}{\linewidth}
\centering
\begin{tabularx}{\textwidth}{c*{4}{C}}
\toprule
          & Number of Reports & Number of categories & Avg. length (char) & Stdev. length (char) \\ \midrule
Dataset 1 & 82,281         & 20            & 410                       & 321                         \\ \midrule
Dataset 2 & 1,625,512       & 9             & 327                       & 174                         \\ \bottomrule
\end{tabularx}

\end{subtable}
\end{table}

\begin{table}[t]
  \scriptsize
      \setlength{\tabcolsep}{4.3pt}
  \caption{Categories in the larger dataset (Dataset 2)}
    \label{tab:data2}
\centering
\begin{tabular}{lclc}
\toprule
Category                                  & Count & Category & Count \\ \midrule
Procedure/Treatment/Test Error & 370K  & Miscellaneous                             & 140K  \\
Medication Error                          & 135K  & Adverse Drug Reaction                     & 34K   \\
Fall                                      & 242K  & Equipment/Supplies/Devices                & 34K   \\
Procedure/Treatment/Test Complication  & 233K  & Transfusion                               & 23K \\
Skin Integrity              & 234K & & \\   \bottomrule
\end{tabular}
\end{table}

%
% ---- Experiments
%
\section{Experiments}

\begin{table}[t]
\centering
\caption{\footnotesize Results on the each dataset on both the validation and test sets. Numbers are percentages. Last row shows our method. $\dagger$ ($\ddagger$) shows statistically significant improvement (McNemar's test) over the next best performance with $p<$0.05 ($p<$0.01).\vspace{6pt}}\label{tab:res}
\setlength{\tabcolsep}{2.7pt}
\renewcommand{\arraystretch}{0.75}
\scriptsize
\begin{tabular}{lcccccccc}
\toprule
\multicolumn{1}{l}{\multirow{3}{*}{Methods}} & \multicolumn{4}{c}{Dataset 1}                       & \multicolumn{4}{c}{Dataset 2}                       \\ \cmidrule{2-9}
\multicolumn{1}{c}{}                         & \multicolumn{2}{c}{Val} & \multicolumn{2}{c}{Test} & \multicolumn{2}{c}{Val} & \multicolumn{2}{c}{Test} \\ \cmidrule{2-9}
\multicolumn{1}{c}{}                         & Acc         & F1         & Acc         & F1          & Acc         & F1         & Acc         & F1          \\ \midrule
SVM \cite{wang2012baselines}                   & 70.7        & 70.3       & 70.9        & 70.6        & 84.8        & 84.0       & 84.7        & 83.9        \\
MNB \cite{wang2012baselines}                   & 71.2        & 71.5       & 71.0        & 72.3        & 79.2        & 79.9       & 79.0        & 79.6        \\
XGB \cite{Chen2016xgb}                         & 71.4        & 69.9       & 72.1        & 70.8        & 76.8        & 75.7       & 76.7        & 75.5        \\ \midrule
cBoW \cite{Zhao2015}                           & 67.5        & 62.6       & 68.0        & 63.4        & 84.8        & 84.2       & 84.6        & 84.1        \\
Adaptive cBoW \cite{Zhao2015}                  & 69.2        & 63.4       & 70.6        & 69.6        & 83.9        & 84.3       & 84.8        & 84.8       \\ \midrule
CNN \cite{kim:2014}                            & 73.2        & 70.7       & 72.2        & 69.5        & 83.6        & 83.1       & 82.7        & 83.5        \\
RNN \cite{dai2015semi}                                           & 76.0        & 74.6       & 74.5        & 72.9        & 84.0        & 84.2       & 83.8        & 83.2        \\
Bi-RNN  \cite{dai2015semi}                                        & 76.3        & 74.5       & 75.2        & 73.6        & 84.7        & 84.3       & 84.6        & 84.5        \\
CNN-BiRNN \cite{tang2015document}             & 77.8        & 76.9       & 76.6        & 76.4        & \bf{89.3}        & 85.9      & 86.8        & 84.6        \\ \midrule
Att-CNN-BiRNN (ours)                        & \bf{78.3} $\dagger$        & \bf{77.2}       & \bf{78.1}$\ddagger$        & \bf{77.3} $\ddagger$        & 89.1        & \bf{88.1}$\ddagger$       & \bf{88.9}$\ddagger$        & \bf{88.0}$\ddagger$        \\ \bottomrule
\end{tabular}
\end{table}

\textbf{Setup. }
We evaluate the effectiveness of our model on two large scale patient safety data obtained from a large healthcare providers in mid-Atlantic US and the Institute for Safe Medication Practices (ISMP). ISMP serves as a safe harbor for all PSE reports from hospitals in Pennsylvania, US. The dataset that was analyzed contains all categories of safety reports (fall, medication, surgery, etc.) and is not limited to medication reports. This study was approved by the MedStar Health Research Institute Institutional Review Board (protocol 2014-101). The characteristics of the data and the categories are shown in tables \ref{tab:data1} and \ref{tab:data2}. We split the data with stratified random sampling into 3 sets: train, validation, and test. We tune the parameters of the neural models on the validation set and the test set remains unseen to the models. We compare our results with conventional text classification models (bag of words feature representation with different types of classifiers), as well as related work on neural architectures (CNNs, RNNs and Bidirectional RNNs and their combinations). For space limitation, we do not explain the details of the baselines and refer the reader to the corresponding citations in Table \ref{tab:res}. We report accuracy and average F1-score results for the categories which are standard evaluation metrics for this task.

\textbf{Implementation. }
We used Keras and TensorFlow for the implementation. We empirically made the following design choices:
We used Word2Vec \cite{mikolov2013distributed} for training the embeddings on both general (Wikipedia) and domain specific corpora (PubMed), similar to \cite{soldaini2017learning}.
We used dropout rates of 0.25 for the recurrent and 0.5 for the convolutional layers. Training was done using Adam optimizer with categorical cross entropy loss; we also applied early stopping for the training.
Number of epochs for the larger dataset was 2 and for small dataset 6.
%Each epoch on an Nvidia GTX 960 GPU for the small and large dataset took about 20 minutes and 2.5 hours, respectively.

\textbf{Results. }
Table \ref{tab:res} demonstrates our main results. As illustrated, our method (last row) significantly outperforms all other methods in virtually all the datasets. This shows the general effectiveness of our model in comparison with the prior work. We observe that our method's performance improvement is slightly larger in the second (larger) dataset. This is expected since our model can better learn the parameters when trained on larger data. Improvement over RNN and CNN-Bi-RNN baselines shows the effectiveness of the neural soft attention in capturing salient parts of the sequence in comparison with the models without attention.

\begin{figure}[b]
\centering
{\includegraphics[width=1.2in]{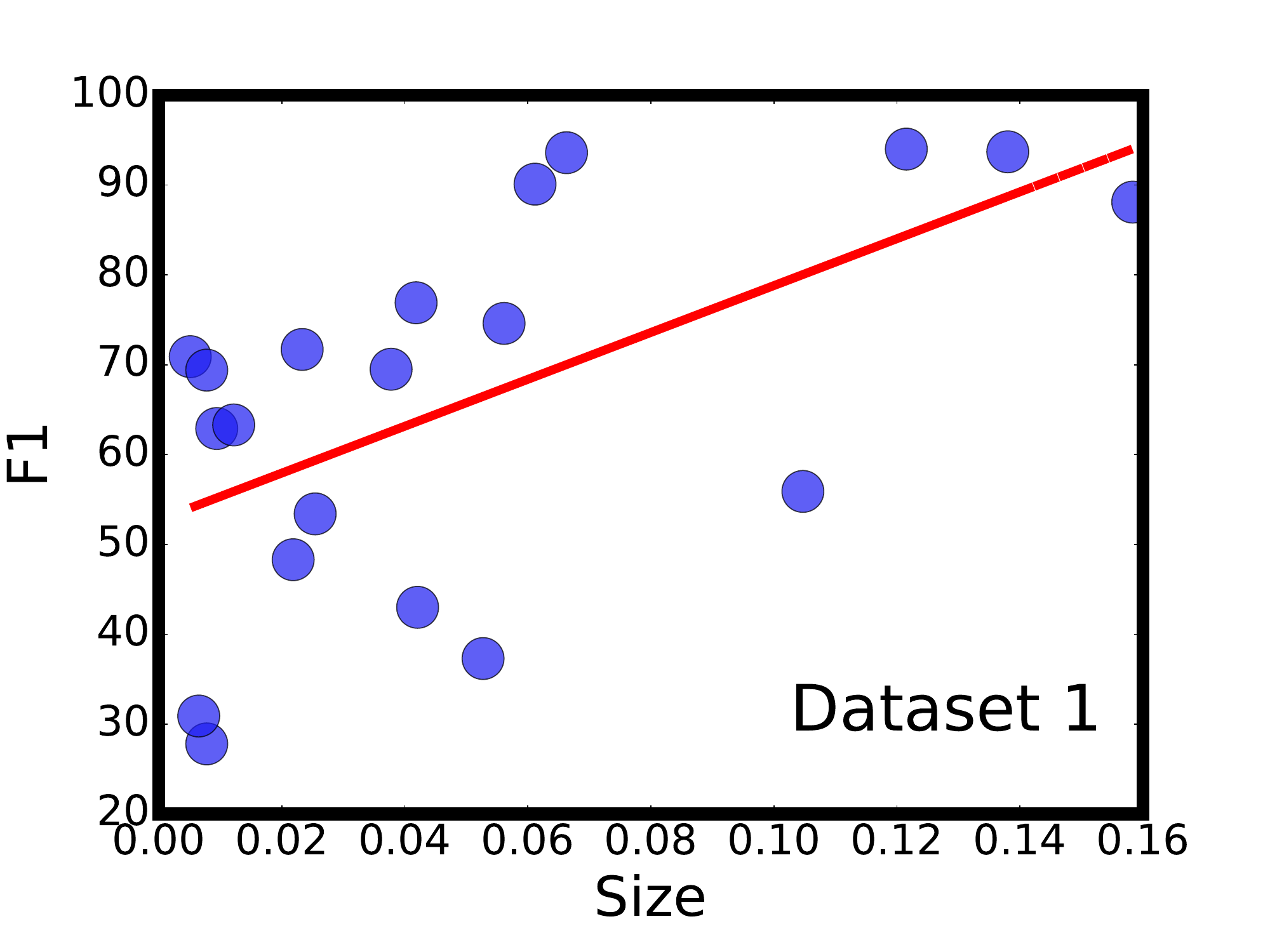}}\hspace{5em}%
{\includegraphics[width=1.2in]{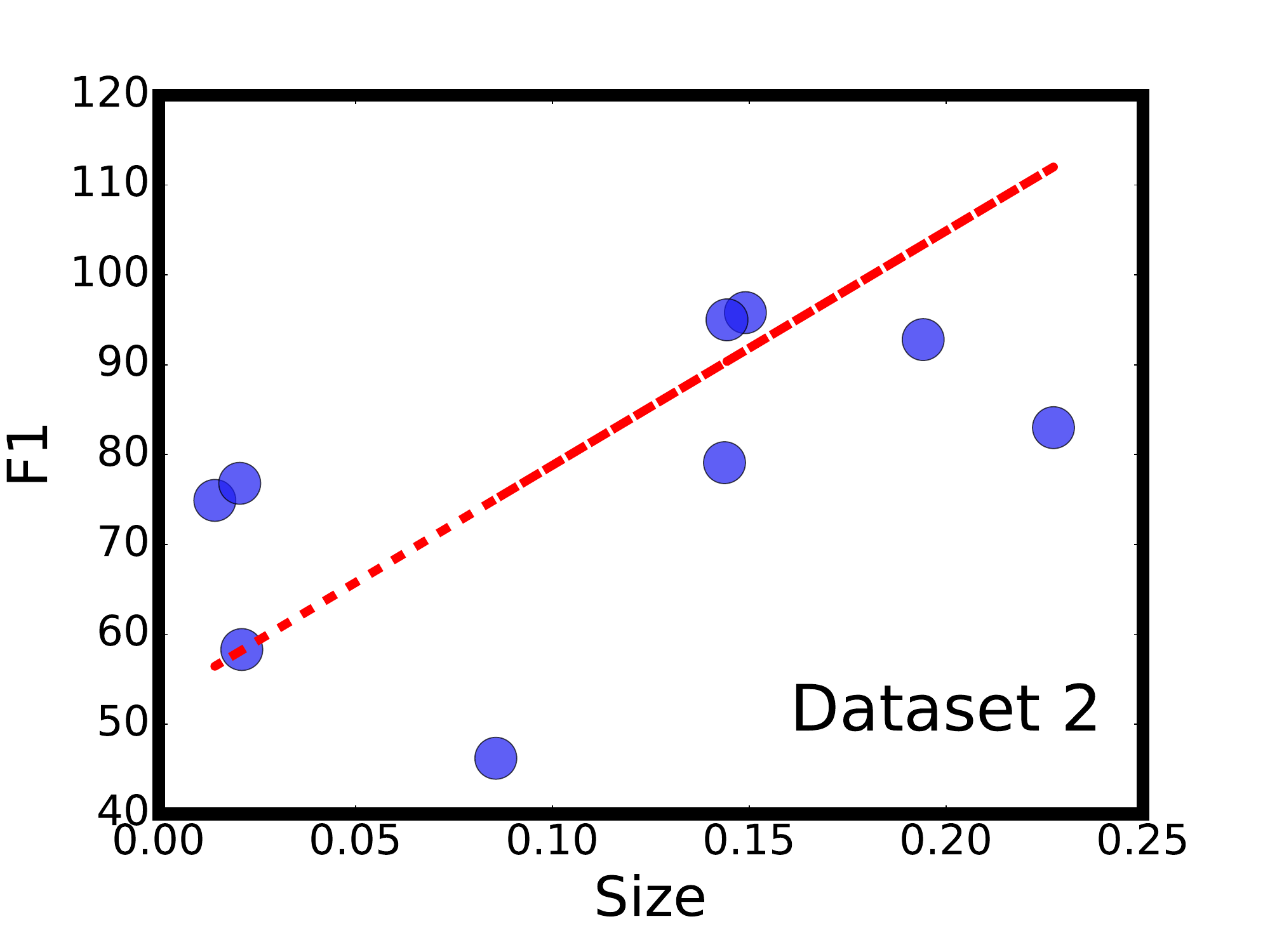}}
\caption{Performance for each category based on its relative size to the dataset.}
\label{fig:test}

\end{figure}

\textbf{Error Analysis. }
While our method effectively outperforms the prior work, we conducted error analysis to better understand the cases that our method fails to correctly perform categorization. In particular, we observed that for both datasets, many wrongly classified samples in the categories were misclassified as the \textsc{``miscellaneous''} category. This pattern was more common for the categories with smaller training samples. This shows that the model learns a broader set of texts for the \textsc{``miscellaneous''} category, which is expected, given the broad nature of this category. We also observed some misclassified samples in the categories that are closely related together. For example in dataset 1, 32\% of the misclassified samples in the \textsc{``blood-bank''} category were classified as \textsc{``lab/specimen''}. A similar pattern was observed for the \textsc{``diagnosis''} and \textsc{``medication''} safety events. These closely related categories usually have overlaps in terms of training data and this makes it hard for the model to differentiate the edge cases. We furthermore observe that the performance on each category correlates with the number of samples in that category. Figure \ref{fig:test} shows this correlation. We observe that generally, our method performs better with the categories of larger relative size. While the correlation is stronger for dataset 1, both datasets show similar trends. This shows that having more training samples helps our model in better learning the characteristics of that particular category and results in higher performance.

%
% ---- Conclusion ----
%
\section{Conclusion}
%Categorizing patient safety events is on of initial steps in preventing medical errors in healthcare centers.
We presented a neural network model based on a soft attention mechanism for categorizing patient safety event reports. We demonstrated the effectiveness of our model on two large-scale real-world datasets and we obtained significant improvements over existing methods. The impact of our method and results is substantial on the patient safety and healthcare, as better categorization of events results in better patient management and prevention of harm to the individuals.

%
% ---- Acknowledgments ----
%
\section*{Acknowledgments}
We thank the 3 anonymous reviewers for their helpful comments. This project was funded under contract/grant number Grant R01 HS023701-02 from the Agency for Healthcare Research and Quality (AHRQ), U.S. Department of Health and Human Services. The opinions expressed in this document are those of the authors and do not necessarily reflect the official position of AHRQ or the U.S. Department of Health and Human Services.

\bibliographystyle{splncs03}
% \bibliography{llncs}
\bibliography{refs-full}
\clearpage

% \addtocmark[2]{Author Index} % additional numbered TOC entry
% \renewcommand{\indexname}{Author Index}
% \printindex
% \clearpage
% \addtocmark[2]{Subject Index} % additional numbered TOC entry
% \markboth{Subject Index}{Subject Index}
% \renewcommand{\indexname}{Subject Index}
% \input{subjidx.ind}
\end{document}